\documentclass[a4paper,11pt]{article}
\usepackage[margin=2cm]{geometry}
\usepackage{amsfonts}
\usepackage{amsmath}
\usepackage{mathptmx}
\usepackage{amssymb}
\usepackage{graphics}
\usepackage{graphicx}
\usepackage{times}
\usepackage[linktocpage,colorlinks=true,urlcolor=blue,citecolor=red,bookmarks=false]{hyperref}
\usepackage[font=small,bf]{caption}
\usepackage{morefloats}
\usepackage{lscape}
\usepackage{listings}

\usepackage{color}

\definecolor{mygreen}{rgb}{0,0.6,0}
\definecolor{mygray}{rgb}{0.5,0.5,0.5}
\definecolor{mymauve}{rgb}{0.58,0,0.82}

\title{InfiniTAM v3: \\ A Framework for Large-Scale 3D Reconstruction with Loop Closure}
\author{
Victor Adrian Prisacariu  \\ \url{victor@robots.ox.ac.uk}  \\ University of Oxford \and
Olaf~K\"{a}hler \\ \url{olaf@robots.ox.ac.uk} \\ University of Oxford \and
Stuart Golodetz \\ \url{smg@robots.ox.ac.uk} \\ University of Oxford \and
Michael Sapienza \\ \url{mikesapi@gmail.com} \\ Samsung Research America \and
Tommaso Cavallari \\ \url{tommaso@robots.ox.ac.uk} \\ University of Oxford \and
Philip H S Torr \\ \url{phst@robots.ox.ac.uk} \\ University of Oxford \and
David W Murray \\ \url{dwm@robots.ox.ac.uk} \\ University of Oxford}

\begin{document}
\maketitle

\tableofcontents

\section{Introduction} \label{s:infinitam_introduction}

This report describes the technical implementation details of InfiniTAM v3, the third version of our InfiniTAM system. It is aimed at closing the gap between the theory behind KinectFusion systems such as~\cite{neissner_tog_2013,newcombe_ismar_2011} and the actual software implementation in our InfiniTAM package.

\subsection{What's New}

In comparison to previous versions of InfiniTAM, we have added several exciting new features, as well as making numerous enhancements to the low-level code that significantly improve our camera tracking performance. The new features that we expect to be of most interest are:
\begin{enumerate}
\item A robust camera tracking module.
We built on top of the tracker released in InfiniTAM v2 and made the system more robust.
The new tracker allows the estimation of camera pose via the alignment of depth images with a raycast of the scene (as before) as well as with the alignment of consecutive RGB frames, to better estimate the camera pose in the presence of geometrically poor surfaces.
Additionally, we implemented a tracking quality evaluation system allowing the detection of tracking failures so as to trigger the relocalisation system described next.

\item An implementation \cite{kahler_eccv_2016} of Glocker et al.'s keyframe-based random ferns camera relocaliser \cite{glocker_tvcg_2015}. This allows for recovery of the camera pose when tracking fails.

\item A novel approach to globally-consistent TSDF-based reconstruction \cite{kahler_eccv_2016}, based on dividing the scene into rigid submaps and optimising the relative poses between them (based on accumulated inter-submap constraints) to construct a consistent overall map.

\item An implementation of Keller et al.'s surfel-based reconstruction approach \cite{keller_3dtv_2013}. Surfels are an interesting alternative to TSDFs for scene reconstruction. On the one hand, they can be useful for handling dynamic scenes, since the surface of a surfel model can be moved around without also needing to update a truncation region around it. On the other hand, visibility determination and collision detection in surfel scenes can be significantly more costly than when using a TSDF. The implementation we have added to InfiniTAM is designed to make it easy for others to explore this trade-off within a single framework.
\end{enumerate}
Our camera tracking extensions are described in Subsection~\ref{s:tracking}. We describe all of the other new features in more detail in Section~\ref{s:newfeatures}.

\section{Cross-Device Implementation Architecture} \label{s:itm_codearch}
\begin{figure}[htpb]
\centering
\includegraphics[width=0.85\linewidth]{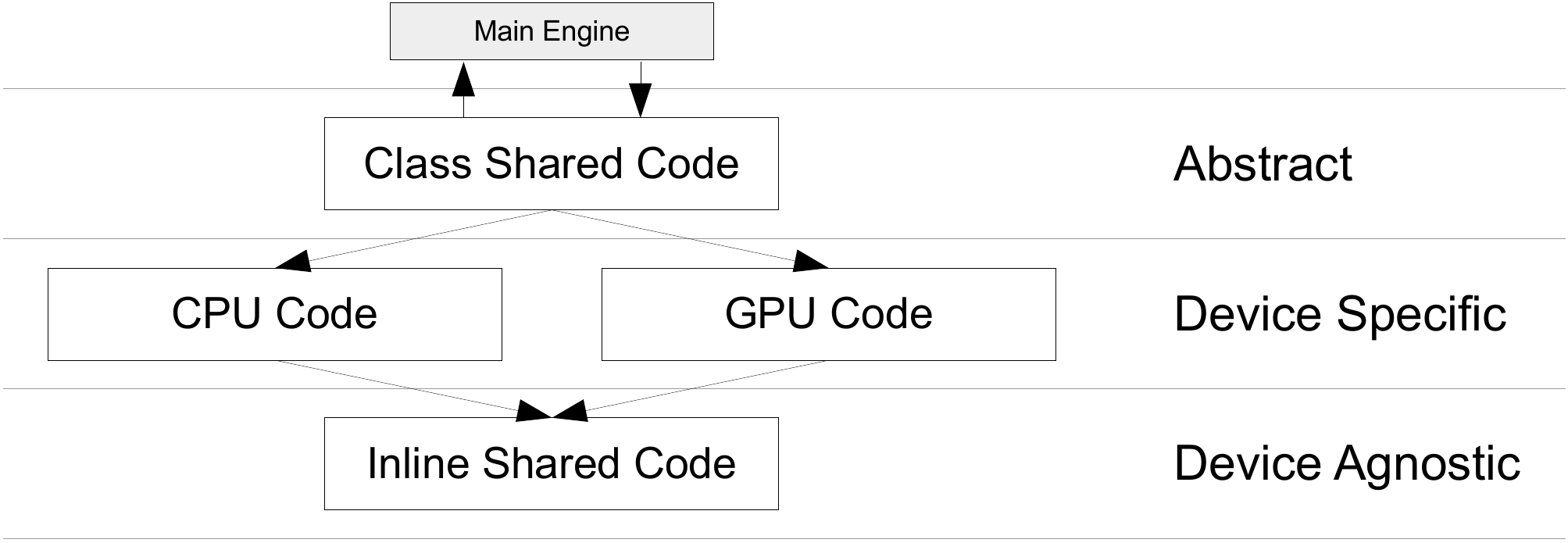}
\caption{InfiniTAM Cross-Device Engine Architecture}
\label{fig:itm_enginestruct}
\end{figure} 

Our implementation follows the chain-of-responsibility design pattern, where
data structures (e.g. ITMImage) are passed between several processing engines
(e.g. ITMTrackingEngine). The engines are ideally stateless and each engine is
responsible for one specific aspect of the overall processing pipeline.
The state is passed on in objects containing the processed information.
Finally one parent class (ITMMainEngine) holds instances of all objects and
engines controls the flow of information.

Each engine is further split into 3 layers as shown in
Figure~\ref{fig:itm_enginestruct}. The topmost, so called \textit{Abstract Layer}
is accessed by the library's main engine and is in general just an abstract
interface, although some common code may be shared at this point. The abstract
interface is implemented in the next, \textit{Device Specific Layer}, which may
be very different between e.g. a CPU and a GPU implementation. Further
implementations using e.g. OpenMP or other hardware acceleration architectures
are possible. At the third, \textit{Device Agnostic Layer}, there is some
inline C-code that may be called from the higher layers.

For the example of a tracking engine, the \textit{Abstract Layer} could contain
code for generic optimization of an error function, the \textit{Device Specific
Layer} could contain a loop or CUDA kernel call to evaluate the error function
for all pixels in an image, and the \textit{Device Agnostic Layer} contains a
simple inline C-function to evaluate the error in a single pixel.

Note that in InfiniTAM v3 the source code files have changed location. Whereas in InfiniTAM v2 there was a single folder containing Device Agnostic and Device Specific files, we now provide separate folders for each module (Tracking, Visualisation, etc.), each split into abstract, specific and agnostic. 

\section{Volumetric Representation with Hashes}\label{s:itm_data_structures}
The key component allowing InfiniTAM to scale KinectFusion to large scale 3D
environments is the volumetric representation using a hash lookup, as also
presented in~\cite{neissner_tog_2013}. This has remained mostly unchanged from InfiniTAM v2.

The data structures and corresponding operations to implement this representation are as follows:
\begin{itemize}
\item Voxel Block Array: Holds the fused color and 3D depth information -- details in Subsection \ref{ss:itm_vba}.
\item Hash Table and Hashing Function: Enable fast access to the voxel block array -- details in Subsection \ref{ss:itm_hashtable}.
\item Hash Table Operations: Insertion, retrieval and deletion -- details in Subsection \ref{ss:itm_hashopers}.
\end{itemize}

\subsection{The Voxel Block Array} \label{ss:itm_vba}
Depth and (optionally) colour information is kept inside an \textbf{ITMVoxel\_[s,f][~, \_rgb, \_conf]} object:
\begin{lstlisting}
struct ITMVoxel_type_variant
{
	/** Value of the truncated signed distance transformation. */
	SDF_DATA_TYPE sdf;
	/** Number of fused observations that make up @p sdf. */
	uchar w_depth;
	/** RGB colour information stored for this voxel. */
	Vector3u clr;
	/** Number of observations that made up @p clr. */
	uchar w_color;

	_CPU_AND_GPU_CODE_ ITMVoxel()
	{
		sdf = SDF_INITIAL_VALUE;
		w_depth = 0;
		clr = (uchar)0;
		w_color = 0;
	}
};
\end{lstlisting}
where (i) \texttt{SDF\_DATA\_TYPE} and \texttt{SDF\_INITIAL\_VALUE} depend on the type of depth representation, which in the current implementation is either float, with initial value 1.0 (\textbf{ITMVoxel\_f\ldots}), or short, with initial value 32767 (\textbf{ITMVoxel\_s\ldots}), (ii) voxel types may or may not store colour information (\textbf{ITMVoxel\_f\_rgb} and \textbf{ITMVoxel\_s\_rgb}) or confidence information (\textbf{ITMVoxel\_f\_conf}), and (iii) \texttt{\_CPU\_AND\_GPU\_CODE\_} defines methods and functions that can be run both as host and as device code:
\begin{lstlisting}
#if defined(__CUDACC__) && defined(__CUDA_ARCH__)
#define _CPU_AND_GPU_CODE_ __device__	// for CUDA device code
#else
#define _CPU_AND_GPU_CODE_ 
#endif
\end{lstlisting}

Voxels are grouped in blocks of predefined size (currently defined as $8\times8\times8$). All the blocks are stored as a contiguous array, referred henceforth as the \textit{voxel block array} or VBA. In the implementation this has a defined size of $2^{18}$ elements. 

\subsection{The Hash Table and The Hashing Function} \label{ss:itm_hashtable}
\begin{figure}[htpb]
\centering
\includegraphics[width=0.85\linewidth]{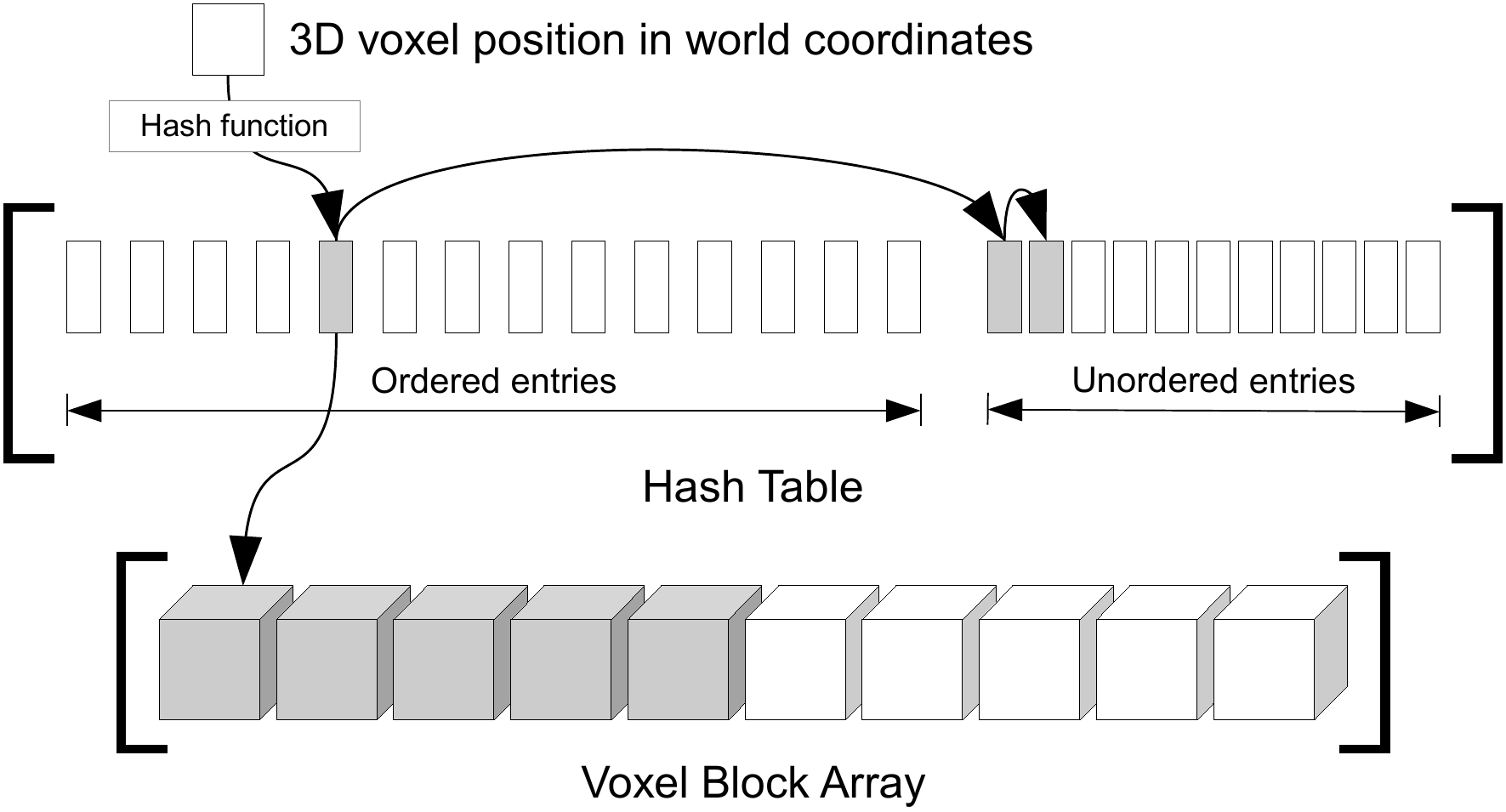}
\caption{Hash table logical structure}
\label{fig:itm_datastruct}
\end{figure}

To quickly and efficiently find the position of a certain voxel block in the
voxel block array, we use a hash table. This hash table is a contiguous array
of \textbf{ITMHashEntry} objects of the following form:
\begin{lstlisting}
struct ITMHashEntry
{
	/** Position of the corner of the 8x8x8 volume, that identifies the entry. */
	Vector3s pos;
	/** Offset in the excess list. */
	int offset;
	/** Pointer to the voxel block array.
	    - >= 0 identifies an actual allocated entry in the voxel block array
	    - -1 identifies an entry that has been removed (swapped out)
	    - <-1 identifies an unallocated block
	*/
	int ptr;
};
\end{lstlisting}
The hash function \texttt{hashIndex} for locating entries of the hash table
takes the corner coordinates \texttt{blockPos} of a 3D voxel block and computes
an index as follows~\cite{neissner_tog_2013}:
\begin{lstlisting}
template<typename T> _CPU_AND_GPU_CODE_ inline int hashIndex(const THREADPTR(T) & blockPos) 
{
	return (((uint)blockPos.x * 73856093u) ^ ((uint)blockPos.y * 19349669u) ^ ((uint)blockPos.z * 83492791u)) & (uint)SDF_HASH_MASK;
}
\end{lstlisting}
To deal with \textit{hash collisions}, each hash index points to a
\textit{bucket} of size 1, which we consider the
\textit{ordered} part of the hash table. There is an additional
\textit{unordered} excess list that is used once an ordered bucket fills up.
In either case, each \textbf{ITMHashEntry} in the hash table stores an offset
in the voxel block array and can hence be used to localize the voxel data for
each specific voxel block. This overall structure is illustrated in
Figure~\ref{fig:itm_datastruct}.

\subsection{Hash Table Operations} \label{ss:itm_hashopers}
The three main operations used when working with a hash table are the \textbf{insertion}, \textbf{retrieval} and \textbf{removal} of entries. In the current version of InfiniTAM we support the former two, with \textbf{removal} not currently required or implemented. The code used by the \textbf{retrieval} operation is shown below:
\begin{lstlisting}
template<class TVoxel>
_CPU_AND_GPU_CODE_ inline TVoxel readVoxel(const CONSTPTR(TVoxel) *voxelData, const CONSTPTR(ITMLib::ITMVoxelBlockHash::IndexData) *voxelIndex, const THREADPTR(Vector3i) & point, THREADPTR(int) &vmIndex, THREADPTR(ITMLib::ITMVoxelBlockHash::IndexCache) & cache)
{
	Vector3i blockPos;
	int linearIdx = pointToVoxelBlockPos(point, blockPos);

	if IS_EQUAL3(blockPos, cache.blockPos)
	{
		vmIndex = true;
		return voxelData[cache.blockPtr + linearIdx];
	}

	int hashIdx = hashIndex(blockPos);

	while (true)
	{
		ITMHashEntry hashEntry = voxelIndex[hashIdx];

		if (IS_EQUAL3(hashEntry.pos, blockPos) && hashEntry.ptr >= 0)
		{
			cache.blockPos = blockPos; cache.blockPtr = hashEntry.ptr * SDF_BLOCK_SIZE3;
			vmIndex = hashIdx + 1; // add 1 to support legacy true / false operations for isFound
	
			return voxelData[cache.blockPtr + linearIdx];
		}
	
		if (hashEntry.offset < 1) break;
		hashIdx = SDF_BUCKET_NUM + hashEntry.offset - 1;
	}

	vmIndex = false;
	return TVoxel();
}
\end{lstlisting}

Both \textbf{insertion} and \textbf{retrieval} work by enumerating the elements of the linked list stored within the hash table. Given a target 3D voxel location in world coordinates, we first compute its corresponding voxel block location, by dividing the voxel location by the size of the voxel block array. Next, we call the hashing function \texttt{hashIndex} to compute the index of the bucket from the ordered part of the hash table. All elements in the bucket are then checked, with \textbf{retrieval} looking for the target block location and \textbf{insertion} for an unallocated hash entry. If this is found, \textbf{retrieval} returns the voxel stored at the target location within the block addressed by the hash entry. \textbf{Insertion} (i) reserves a block inside the voxel block array and (ii) populates the hash table with a new entry containing the reserved voxel block array address and target block 3D world coordinate location.

If all locations in the bucket are exhausted, the enumeration of the linked list moves to the unordered part of the hash table, using the \textit{offset} field to provide the location of the next hash entry. The enumeration finishes when \textit{offset} is found to be smaller or equal to $-1$. At this point, if the target location still has not been found, \textbf{retrieval} returns an empty voxel. \textbf{Insertion} (i) reserves an unallocated entry in the unordered part of the hash table and a block inside the voxel block array, (ii) populates the hash table with a new entry containing the reserved voxel block array address and target block 3D world coordinate location and (iii) changes the \textit{offset} field in the last found entry in the linked list to point to the newly populated one.

The reserve operations used for the unordered part of the hash table and for the voxel block array use prepopulated allocation lists and, in the GPU code, atomic operations. 

All hash table operations are done through these functions and there is no direct memory access encouraged or indeed permitted by the current version of the code. 

\section{Method Stages}\label{s:itm_methodstages}
\begin{figure}[htpb]
\centering
\includegraphics[width=\linewidth]{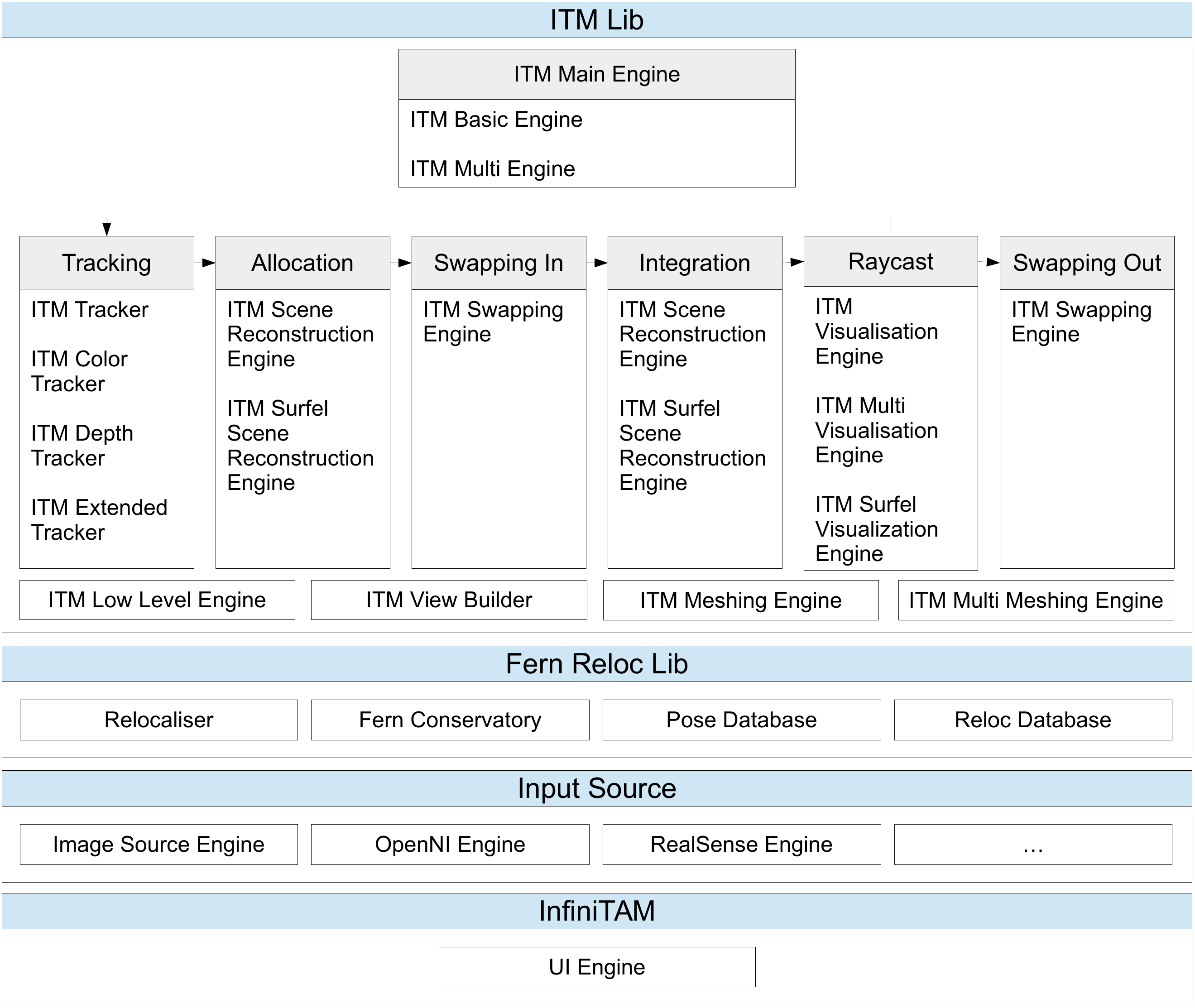}
\caption{InfiniTAM logical and technical overview.}
\label{fig:itm_method_overview}
\end{figure} 

Each incoming frame from the camera is processed by a pipeline similar to the ones in~\cite{neissner_tog_2013} and the original KinectFusion system~\cite{newcombe_ismar_2011}, as shown in Figure \ref{fig:itm_method_overview}.
\texttt{ITMMainEngine} is the abstract class containing the entry point to the pipeline, in form of the \texttt{ProcessFrame} method.
There are two concrete pipeline implementations: \texttt{ITMBasicEngine}, a standard fusion one, similar to that described in~\cite{newcombe_ismar_2011,neissner_tog_2013}, and \texttt{ITMMultiEngine} allowing the generation of \emph{globally-consistent reconstructions} (new feature in InfiniTAM v3), as described in Section~\ref{s:loopclosure}.
Each call to the \texttt{ProcessFrame} method of a pipeline can be logically split in six distinct stages, each implemented using one or more processing engines:
\begin{itemize}
\item \textbf{Tracking}: The camera pose for the new frame is obtained by fitting the current depth (and optionally color) image to the projection of the world model from the previous frame. This is implemented using the \texttt{ITMTracker}, \texttt{ITMColorTracker}, \texttt{ITMDepthTracker}, and \texttt{ITMExtendedTracker} engines.
\item \textbf{Allocation}: Based on the depth image, new voxel blocks are allocated as required and a list of all visible voxel blocks is built. This is implemented inside the \texttt{ITMSceneReconstructionEngine} (voxel map) and \texttt{ITMSurfelSceneReconstructionEngine} classes (surfel map).
\item \textbf{Swapping In}: If required, map data is swapped in from host memory to device memory. This is implemented using the \texttt{ITMSwappingEngine}. The current version only supports the BasicEngine with voxels.
\item \textbf{Integration}: The current depth and color frames are integrated within the map. Once again, this operation is implemented inside the \texttt{ITMSceneReconstructionEngine} (map represented using voxels) and \texttt{ITMSurfelSceneReconstructionEngine} (surfel map) classes.
\item \textbf{Raycast}: The world model is rendered from the current pose (i) for visualization purposes and (ii) to be used by the tracking stage at the next frame. This uses the \texttt{ITMVisualisationEngine}, \texttt{ITMMultiVisualisationEngine}, and \texttt{ITMSurfelVisualisationEngine} classes.
\item \textbf{Swapping Out}: The parts of the map that are not visible are swapped out from device memory to host memory. This is implemented using the \texttt{ITMSwappingEngine}. The current version only supports the BasicEngine with voxels.
\end{itemize}

The main processing engines are contained within the \texttt{ITMLib} namespace, along with \texttt{ITMLowLevelEngine}, \texttt{ITMViewBuilder}, \texttt{ITMMeshingEngine}, and \texttt{ITMMultiMeshingEngine}. These are used, respectively, for low level processing (e.g. image copy, gradients and rescale), image preparation (converting depth images from \texttt{unsigned short} to \texttt{float} values, normal computation, bilateral filtering), and mesh generation via the Marching-Cubes algorithm~\cite{Lorensen1987}.

The image acquisition routines, relying on a multitude of input sensors, can be found in the \texttt{InputSource} namespace, whilst the relocaliser implementation (another feature added in InfiniTAM v3 and described in Section.~\ref{s:relocaliser}) is contained in the \texttt{FernRelocLib} namespace.

Finally, the main UI and the associated application are contained inside the \texttt{InfiniTAM} namespace.

The following presents a discussion of the tracking, allocation, integration and raycast stages. We delay a discussion of the swapping until Section~\ref{s:swapping}.

\subsection{Tracking}
\label{s:tracking}

In the tracking stage we have to determine the pose of a new image given the
3D world model. We do this either based on the new depth image with an
\texttt{ITMDepthTracker}, or based on the color image with an
\texttt{ITMColorTracker}.
From version 3 of the \texttt{InfiniTAM} system, we also provide a revised and improved tracking algorithm: \texttt{ITMExtendedTracker}.
All extend the abstract \texttt{ITMTracker} class and have device-specific implementations running on the
CPU and on CUDA.

\subsubsection{ITMDepthTracker}
\label{sss:ITMDepthTracker}
In the \texttt{ITMDepthTracker} we follow the original alignment process as
described in~\cite{newcombe_ismar_2011,izadi_uist_2011}:
\begin{itemize}
  \item Render a map $\mathcal{V}$ of surface points and a map $\mathcal{N}$ of
        surface normals from the viewpoint of an initial guess -- details in Section~\ref{s:raycasting}
  \item Project all points $\mathbf{p}$ from the depth image onto points
	$\bar{\mathbf{p}}$ in $\mathcal{V}$ and $\mathcal{N}$ and compute their
        distances from the planar approximation of the surface, i.e.
        $d = \left(\mathbf{R} \mathbf{p} + \mathbf{t} - \mathcal{V}(\bar{\mathbf{p}})\right)^T \mathcal{N}(\bar{\mathbf{p}})$
  \item Find $\mathbf{R}$ and $\mathbf{t}$ minimizing of the sum of the squared
        distances by solving linear equation system
  \item Iterate the previous two steps until convergence
\end{itemize}
A resolution hierarchy of the depth image is used in our implementation to
improve the convergence behaviour.

\subsubsection{ITMColorTracker}
Alternatively the color image can be used within an \texttt{ITMColorTracker}.
In this case the alignment process is as follows:
\begin{itemize}
  \item Create a list $\mathcal{V}$ of surface points and a corresponding list
        $\mathcal{C}$ of colours from the viewpoint of an initial guess --
        details in Section~\ref{s:raycasting}.
  \item Project all points from $\mathcal{V}$ into the current color image $I$
        and compute the difference in colours, i.e.
        $d = \left\|I(\pi(\mathbf{R} \mathcal{V}(i) + \mathbf{t})) - \mathcal{C}(i)\right\|_2$.
  \item Find $\mathbf{R}$ and $\mathbf{t}$ minimizing of the sum of the squared
        differences using the Levenberg-Marquardt optimization algorithm.
\end{itemize}
Again a resolution hierarchy in the color image is used and the list of surface
points is subsampled by a factor of 4.

\subsubsection{ITMExtendedTracker}
The \texttt{ITMExtendedTracker} class allows the alignment of the current RGB-D image captured by the sensor in a manner analogous to that described in Section~\ref{sss:ITMDepthTracker}.

More specifically, in its default configuration, an instance of \texttt{ITMExtendedTracker} can seamlessly replace an \texttt{ITMDepthTracker} whilst providing typically better tracking accuracy.
The main differences with a simple ICP-based tracker (such as the one previously implemented -- and described in~\cite{newcombe_ismar_2011,izadi_uist_2011}) are as follows:
\begin{itemize}
    \item A robust Huber-norm is deployed instead of the quite standard L2 norm when computing the error term associated to each pixel of the input depth image.
    \item The error term for each pixel of the depth image, in addition to being subject to the robust norm just described, is weighted according to its depth measurement provided by the sensor, so as to account for the noisier nature of distance measurements associated to points far away from the camera (weight decreases with the increase in distance reading).
    \item Points $\mathbf{p}$ whose projection $\bar{\mathbf{p}}$ in $\mathcal{V}$ have a distance from the planar approximation of the surface $d = \left(\mathbf{R} \mathbf{p} + \mathbf{t} - \mathcal{V}(\bar{\mathbf{p}})\right)^T \mathcal{N}(\bar{\mathbf{p}})$ greater than a configurable threshold, are not considered in the error function being minimised, as they are likely outliers.
    \item Finally, as described in Section~2.2 of~\cite{kahler_eccv_2016}, the results of the ICP optimisation phase (percentage of inlier pixels, determinant of the Hessian and residual sum) are evaluated by an SVM classifier to separate between tracking success and failure, as well as to establish whether a successful result has good or bad accuracy. In case of tracking failure, the relocaliser described in Section~\ref{s:relocaliser} is activated to attempt the estimation of the current camera pose and recover the localisation and mapping loop.
\end{itemize}

The capabilities of the \texttt{ITMExtendedTracker} class are not limited to a better ICP-based camera tracking step.
The user can, optionally, achieve a combined geometric-photometric alignment as well.

As in~\cite{whelan_ijrr_2014}, we augment the extended tracker's geometric error term described above with a per-pixel photometric error term.
In more detail, we try to minimise the difference in intensity between each pixel in the current RGB image and the corresponding pixel in the previous frame captured by the sensor (note that, differently from the \texttt{ITMColorTracker} described in the previous section, we do not need to have access to a coloured reconstruction, since we do not make use of voxel colours in the matching phase).
Frame-to-frame pixel intensity differences are estimated as follows:
\begin{enumerate}
    \item The intensity and depth value associated to a pixel in the current RGB-D pair are sampled from the images (each intensity value $I$ is computed as a weighted average of the RGB colour channels: $I = 0.299 R + 0.587 G + 0.114 B$).
    \item 3D coordinates of the pixel in the current camera's reference frame are computed by backprojecting its depth value according to the camera intrinsic parameters.
    \item Given a candidate sensor pose (the initial guess also used to compute the geometric error term), the 3D pixel coordinates are brought into a scene coordinate frame.
    \item Use the previous frame's estimated pose to bring the pixel coordinates back to the previous camera's reference frame.
    \item Project those coordinates onto the previous RGB image and compute the intensity of the sampled point by bilinearly interpolating values associated to its neighbouring pixels.
\end{enumerate}

The sum of per-pixel intensity differences over the whole image -- as before, subject to depth-based weighting and robust loss function (we use Tukey's in this case) -- is added to the geometric error term described above using an appropriate scaling factor, to account for the different ranges of the values ($0.3$ in the current implementation).
Gradient and Hessian values are combined as well.
A Levenberg-Marquardt optimization algorithm is then deployed to minimise the error and estimate the final pose.

As with the other trackers, a resolution hierarchy is computed, and the camera alignment is performed starting from the coarsest resolution to the finest one, to improve the convergence behaviour.

\subsubsection{Configuration}
A \texttt{string} in \texttt{ITMLibSettings} allows to select which tracker to use, as well as specifying tracker-specific configuration values.
By default an instance of \texttt{ITMExtendedTracker} (with the colour-tracking energy term disabled -- to allow for the employment of depth-only sensors such as Occipital's Structure\footnote{\url{https://structure.io/}}) is created.
Other configuration strings can be found (as comments) in \texttt{ITMLib/Utils/ITMLibSettings.cpp}.

All tracker implementations use the device (in the CUDA-specific subclass, relying on the CPU otherwise) only for the computation of function, gradient and Hessian values.
Everything else, such as the optimisation, is done in the main, abstract class. 

\subsection{Allocation}
The allocation stage is split into three separate parts. Our aim here was minimise the use of blocking operations (e.g. atomics) and completely avoid the use of critical sections. 

Firstly, for each 2.5D pixel from the depth image, we backproject a line connecting $d-\mu$ to $d+\mu$, where $d$ is the depth in image coordinates and $\mu$ is a fixed, tunable parameter. This leads to a line in world coordinates, which intersects a number of voxel blocks. We search the hash table for each of these blocks and look for a free hash entry for each unallocated one. These memorised for the next stage of the allocation using two arrays, each with the same number of elements as the hash table, containing information about the allocation and visibility of the new hash entries. Note that, if two or more blocks for the same depth image are mapped to the same hash entry (i.e. if we have intra-frame hash collisions), only one will be allocated. This artefact is fixed automatically at the next frame, as the intra-frame camera motion is relatively small. 

Secondly, we allocate voxel blocks for each non zero entry in the allocation and visibility arrays built previously. This is done using atomic subtraction on a stack of free voxel block indices i.e. we decrease the number of available blocks by one and add the previous head of the stack to the hash entry.

Thirdly, we build a list of live hash entries, i.e. containing the ones that project inside the visible frustum. This is later going to be used by the integration and swapping in/out stages. 

\subsection{Integration}

In the integration stage, the information from the most recent images is
incorporated into the 3D world model. This is done essentially the same way as
in the original KinectFusion algorithm~\cite{newcombe_ismar_2011,izadi_uist_2011}. For each voxel in any of
the visible voxel blocks from above the function
\texttt{computeUpdatedVoxelDepthInfo} is called. If a voxel is \textit{behind}
the surface observed in the new depth image, the image does not contain any new
information about it, and the function returns. If the voxel is close to or in
front of the observed surface, a corresponding observation is added to the
accumulated sum. This is illustrated in the listing of the function
\texttt{computeUpdatedVoxelDepthInfo} below.
\begin{lstlisting}
_CPU_AND_GPU_CODE_ inline float computeUpdatedVoxelDepthInfo(ITMVoxel &voxel, Vector4f pt_model, Matrix4f M_d, Vector4f projParams_d,
	float mu, int maxW, float *depth, Vector2i imgSize)
{
	Vector4f pt_camera; Vector2f pt_image;
	float depth_measure, eta, oldF, newF;
	int oldW, newW;

	// project point into image
	pt_camera = M_d * pt_model;
	if (pt_camera.z <= 0) return -1;

	pt_image.x = projParams_d.x * pt_camera.x / pt_camera.z + projParams_d.z;
	pt_image.y = projParams_d.y * pt_camera.y / pt_camera.z + projParams_d.w;
	if ((pt_image.x < 1) || (pt_image.x > imgSize.x - 2) || (pt_image.y < 1) || (pt_image.y > imgSize.y - 2)) return - 1;

	// get measured depth from image
	depth_measure = depth[(int)(pt_image.x + 0.5f) + (int)(pt_image.y + 0.5f) * imgSize.x];
	if (depth_measure <= 0.0) return -1;

	// check whether voxel needs updating
	eta = depth_measure - pt_camera.z;
	if (eta < -mu) return eta;

	// compute updated SDF value and reliability
	oldF = SDF_DATA_TYPE_TO_FLOAT(voxel.sdf); oldW = voxel.w_depth;
	newF = MIN(1.0f, eta / mu);
	newW = 1;

	newF = oldW * oldF + newW * newF;
	newW = oldW + newW;
	newF /= newW;
	newW = MIN(newW, maxW);

	// write back
	voxel.sdf = SDF_FLOAT_TO_DATA_TYPE(newF);
	voxel.w_depth = newW;

	return eta;
}
\end{lstlisting}
The main difference to the original KinectFusion integration step is that the
aforementioned update function is not called for all the voxels in a fixed
volume but only for the voxels in the blocks that are currently marked as
visible.

\subsection{Raycast}\label{s:raycasting}
As the last step in the pipeline, a depth image is computed from the updated
3D world model given a camera pose and intrinsics. This is required as input to
the tracking step in the next frame and also for visualization.
The main underlying process is raycasting, i.e. for each pixel in the image a
ray is being cast from the camera up until an intersection with the surface is
found. This essentially means checking the value of the truncated signed
distance function at each voxel along the ray until a zero-crossing is found.

As noted in the original KinectFusion paper~\cite{newcombe_ismar_2011}, the performance of the
raycasting can be improved significantly by taking larger steps along the ray.
The value of the truncated signed distance function can serve as a conservative
guess for the distance to the nearest surface, hence this value can be used as
step size. To additionally handle empty space in the volumetric representation,
where no corresponding voxel block has been allocated yet, we introduce a state
machine with the following states:
\begin{lstlisting}
SEARCH_BLOCK_COARSE,
SEARCH_BLOCK_FINE,
SEARCH_SURFACE,
BEHIND_SURFACE,
WRONG_SIDE
\end{lstlisting}
Starting from \texttt{SEARCH\_BLOCK\_COARSE}, we take steps of the size of each
block, i.e. $8$ voxels, until an actually allocated block is encountered. Once
the ray enters an allocated block, we take a step back and enter state
\texttt{SEARCH\_BLOCK\_FINE}, indicating that the step length is now limited by
the truncation band of the signed distance transform. Once we enter a valid
block and the values in that block indicate we are still in front of the
surface, the state is changed to \texttt{SEARCH\_SURFACE} until a negative
value is read from the signed distance transform, which indicates we are now
in state \texttt{BEHIND\_SURFACE}. This terminates the raycasting iteration and
the exact location of the surface is now found by two trilinear interpolation
steps. The state \texttt{WRONG\_SIDE} is entered if we are searching for a
valid block in state \texttt{SEARCH\_BLOCK\_FINE} and encounter negative SDF
values, indicating we are behind the surface, as soon as we enter a block. In
this case the ray hits the surface from behind for whichever reason, and we do
not want to count the boundary between the unallocated, empty block and the
block with the negative values as an object surface.
The actual implementation of this process can be found in the \texttt{castRay} method of the \texttt{ITMVisualisationEngine\_Shared.h} file where is possible to see how the step length is computed.

Another measure for improving the performance of the raycasting is to select a
plausible search range. From the allocation step we are given a list of visible
voxel blocks, and we can try and render these blocks by forward projection to
give us an idea of the maximum and minimum depth values to expect at each pixel.
Within InfiniTAM this is done using an \texttt{ITMVisualisationEngine}.
A naive implementation on the CPU computes the 2D bounding box of the projection
of each voxel block into the image and fills this area with the maximum and
minimum depth values of the corresponding 3D bounding box of the voxel block,
correctly handling overlapping bounding boxes, of course.

To parallelise this process on the GPU we split it into two steps. First we
project each block down into the image, compute the bounding box, and create a
list of $16\times{}16$ pixel fragments, that are to be filled with specific
minimum and maximum depth values. Apart from a prefix sum to count the number
of fragments, this is trivially parallelisable. Second we go through the list
of fragments and actually render them. Updating the minimum and maximum depth
for a pixel requires atomic operations, but by splitting the process into
fragments we reduce the number of collisions to typically a few hundreds or
thousands of pixels in a $640\times{}480$ image and achieve an efficiently
parallelised overall process.

\section{Swapping}\label{s:swapping}
Voxel hashing enables much larger maps to be created, compared to the standard KinectFusion approach of using dense 3D volumes. Video card memory capacity however is often quite limited. Practically an off-the-shelf video card can hold roughly the map of a single room at 4mm voxel resolution in active memory, even with voxel hashing. This problem can be mitigated using a traditional method from the graphics community, that is also employed in~\cite{neissner_tog_2013}. We only hold the \textit{active} part of the map in video card memory, i.e. only parts that are inside or close to the current view frustum. The remainder of the map is swapped out to host memory and swapped back in as needed. 

We have designed our swapping framework aiming for the following three objectives: (O1) the transfers between host and device should be minimized and have guaranteed maximum bounds, (O2) host processing time should be kept to a minimum and (O3) no assumptions should be made about the type and speed of the host memory, i.e. it could be a hard drive. These objectives lead to the following design considerations:
\begin{itemize}
\item \textbf{O1}: All memory transfers use a host/device buffer of \textit{fixed} user-defined size. 
\item \textbf{O2}: The host map memory is configured as a voxel block array of size equal to the number of spaces in the hash table. Therefore, to check if a hash entry has a corresponding voxel block in the host memory, only the hash table index needs to be transferred and checked. The host does not need to perform any further computations, e.g. as it would have to do if a separate host hash table were used. Furthermore, whenever a voxel block is deallocated from device memory, its corresponding hash entry is not deleted but rather marked as unavailable in device memory, and, implicitly, available in host memory. This (i) helps maintain consistency between device hash table and host voxel block storage and (ii) enables a fast visibility check for the parts of the map stored only in host memory.
\item \textbf{O3}: Global memory speed is a function of the type of storage device used, e.g. faster for RAM and slower for flash or hard drive storage. This means that, for certain configurations, host memory operations can be considerably slower than the device reconstruction. To account for this behaviour and to enable stable tracking, the device is constantly integrating new live depth data even for parts of the scene that are known to have host data that is not yet in device memory. This might mean that, by the time all visible parts of the scene have been swapped into the device memory, some voxel blocks might hold large amounts of new data integrated by the device. We could replace the newly fused data with the old one from the host stored map, but this would mean disregarding perfectly fine map data. Instead, after the swapping operation, we run a secondary integration that fuses the host voxel block data with the newly fused device map.
\end{itemize}

\begin{figure}[htpb]
\centering
\includegraphics[width=0.7\linewidth]{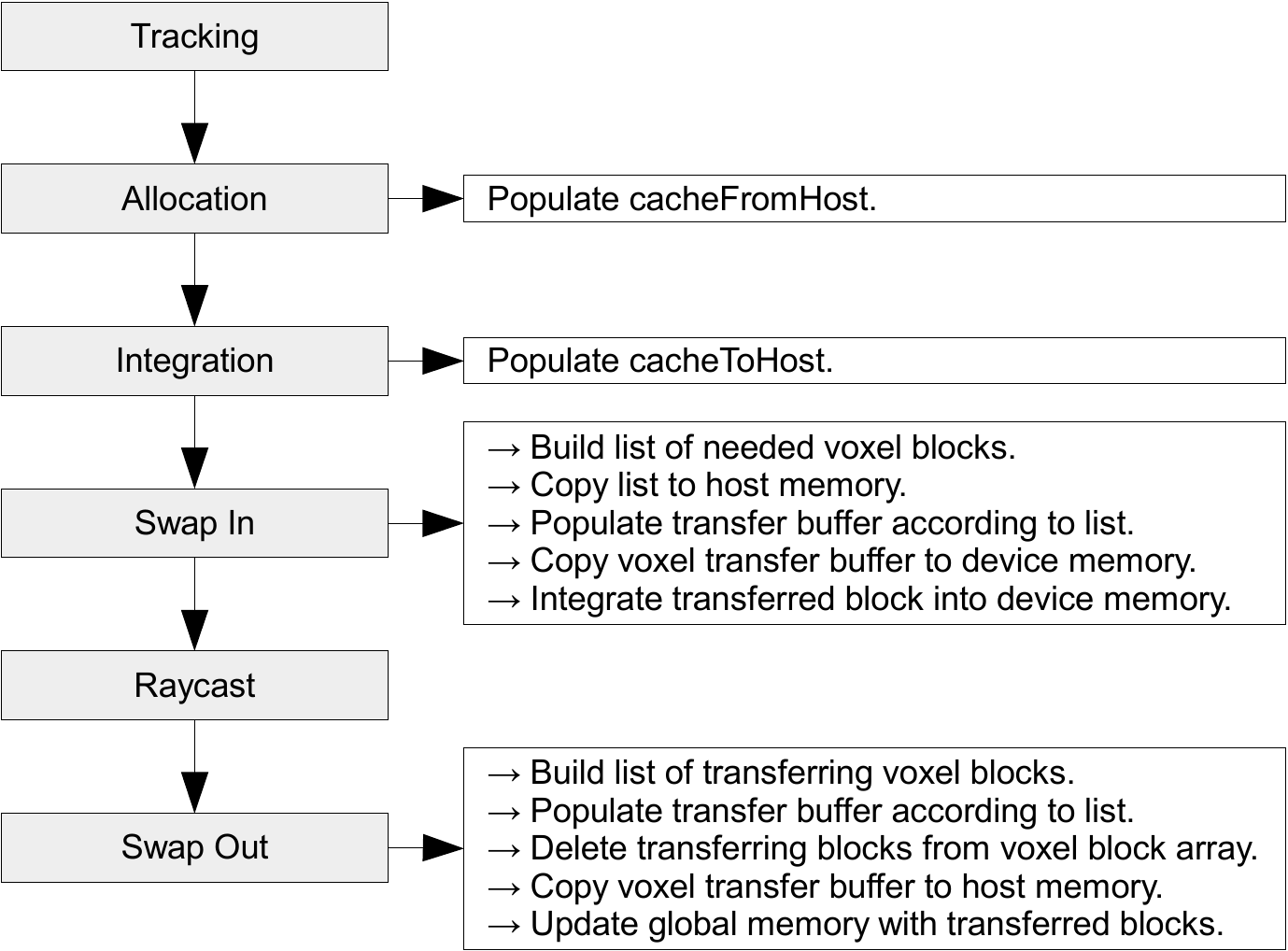}
\caption{Swapping pipeline}
\label{fig:itm_swap_overall}
\end{figure} 

The design considerations have led us to the swapping in/out pipeline shown in Figure \ref{fig:itm_swap_overall}. We use the allocation stage to establish which parts of the map need to be swapped in, and the integration stage to mark which parts need to swapped out. A voxel needs to be swapped (i) from host once it projects within a small (tunable) distance from the boundaries of live visible frame and (ii) to disk after data has been integrated from the depth camera. 

\begin{figure}[htpb]
\centering
\includegraphics[width=0.9\linewidth]{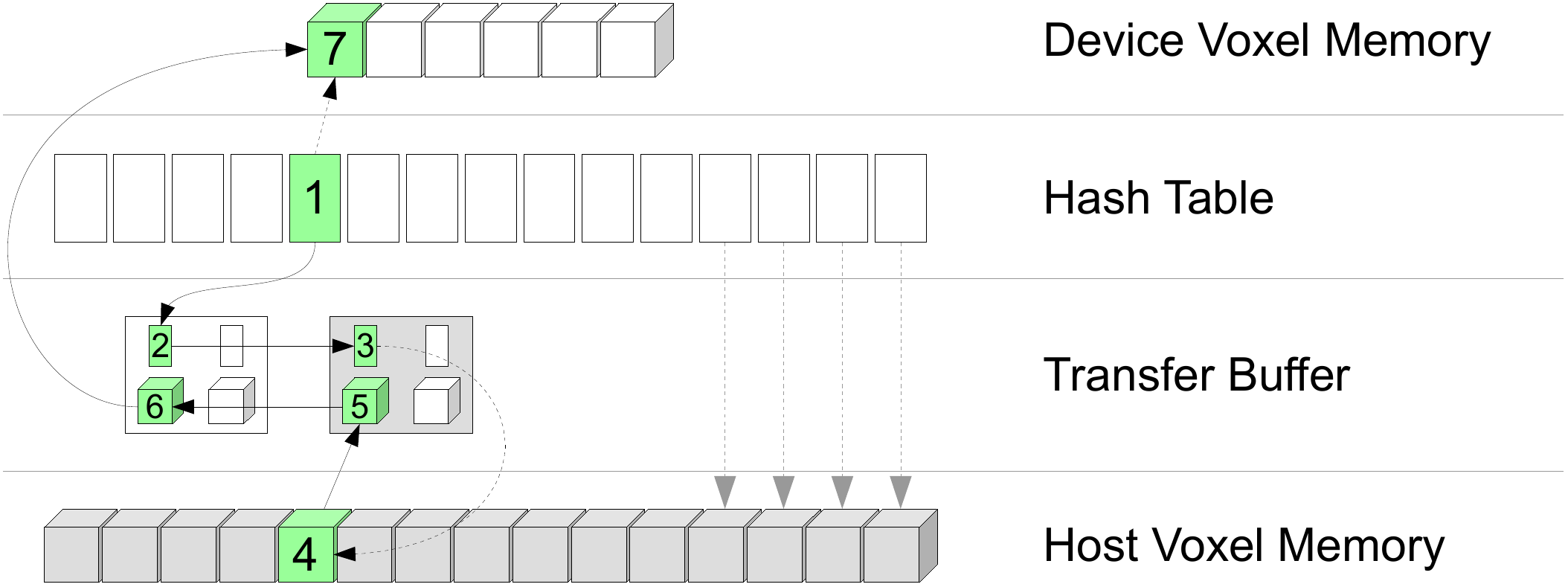}
\caption{Swapping in: First, the hash table entry at address \textbf{1} is copied into the device transfer buffer at address \textbf{2}. This is then copied at address \textbf{3} in the host transfer buffer and used as an address inside the host voxel block array, indicating the block at address \textbf{4}. This block is finally copied back to location \textbf{7} inside the device voxel block array, passing through the host transfer buffer (location \textbf{5}) and the device transfer buffer (location \textbf{6}).}
\label{fig:itm_swap_in}
\end{figure} 

The swapping in stage is exemplified for a single block in Figure \ref{fig:itm_swap_in}. The indices of the hash entries that need to be swapped in are copied into the device transfer buffer, up to its capacity. Next, this is transferred to the host transfer buffer. There the indices are used as addresses inside the host voxel block array and the target blocks are copied to the host transfer buffer. Finally, the host transfer buffer is copied to the device where a single kernel integrates directly from the transfer buffer into the device voxel block memory.

\begin{figure}[htpb]
\centering
\includegraphics[width=0.9\linewidth]{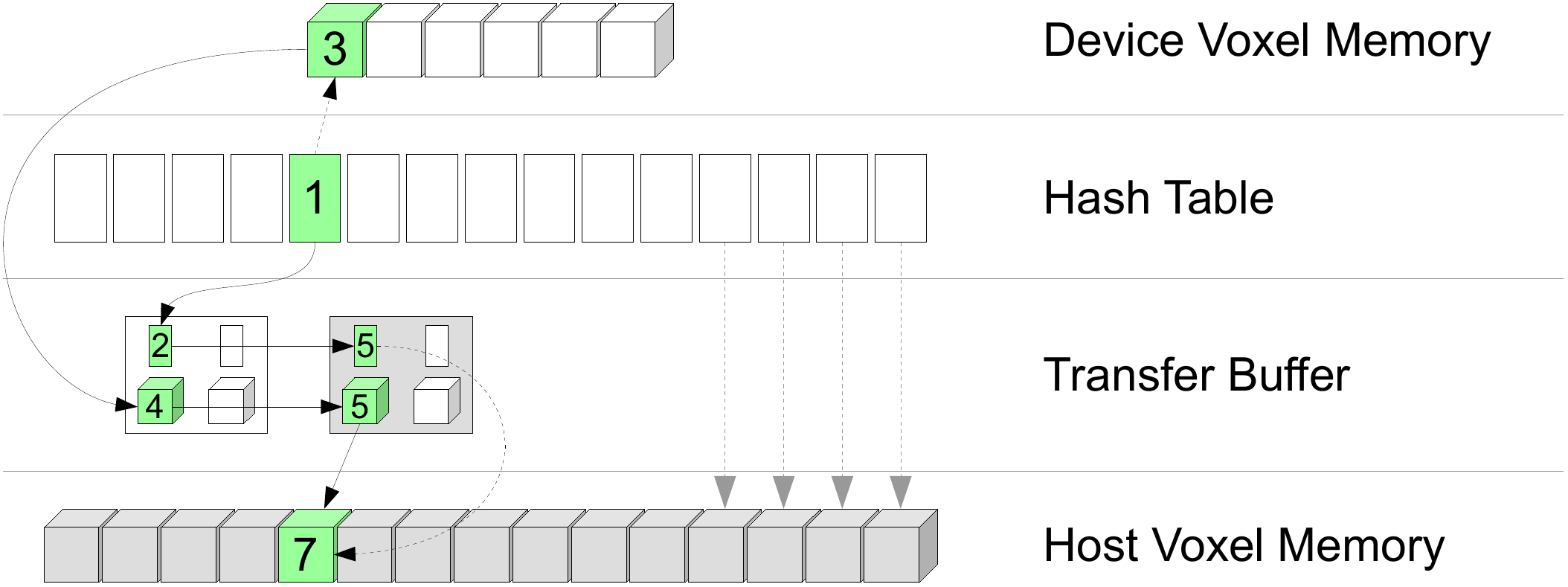}
\caption{Swapping Out. The hash table entry at location \textbf{1} and the voxel block at location \textbf{3} are copied into the device transfer buffer at locations \textbf{2} and \textbf{4}, respectively. The entire used transfer buffer is then copied to the host, at locations \textbf{5}, and the hash entry index is used to copy the voxel block into location \textbf{7} inside the host voxel block array.}
\label{fig:itm_swap_out}
\end{figure} 

An example for the swapping out stage is shown in Figure \ref{fig:itm_swap_out} for a single block. Both indices and voxel blocks that need to be swapped out are copied to the device transfer buffer. This is then copied to the host transfer buffer memory and again to host voxel memory.

All swapping related variables and memory is kept inside the \texttt{ITMGlobalCache} object and all swapping related operations are done by the \texttt{ITMSwappingEngine}. 

\section{UI, Usage and Examples}\label{s:results}
\begin{figure}[htpb]
\centering
\includegraphics[width=0.9\linewidth]{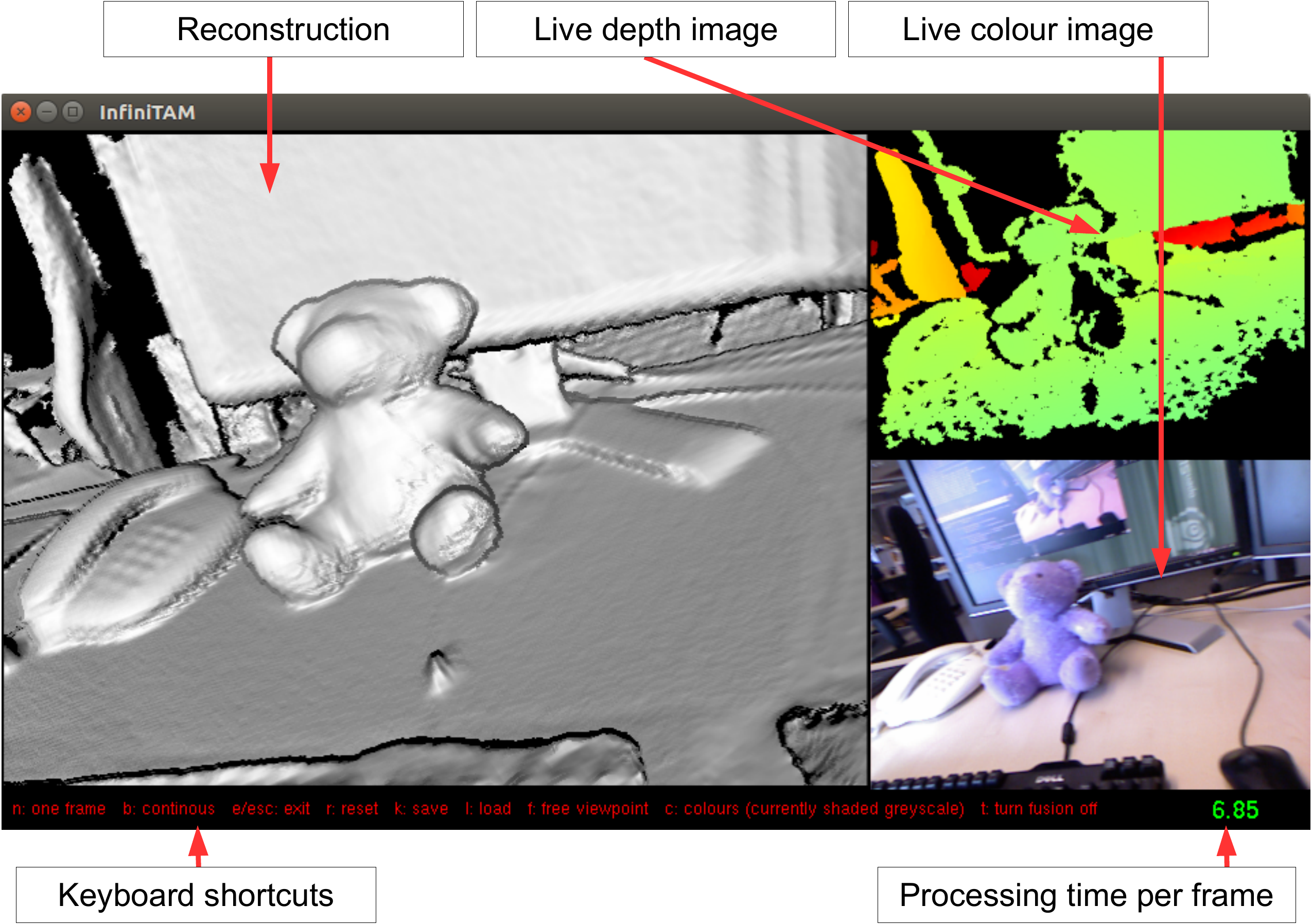}
\caption{InfiniTAM UI Example}
\label{fig:itm_ui}
\end{figure} 

The UI for our implementation is shown in Figure \ref{fig:itm_ui}. We show a live raycasted rendering of the reconstruction, live depth image and live colour image, along with the processing time per frame and the keyboard shortcuts available for the UI. The user can choose to process one frame at a time or process the whole input video stream. Other functionalities such as resetting the InfiniTAM system, exporting a 3D model from the reconstruction, arbitrary view raycast and different rendering styles are also available. The UI window and behaviour is implemented in \texttt{UIEngine} and requires OpenGL and GLUT.

Running InfiniTAM requires a calibrated data source of depth (and optionally image). The calibration is specified through an external file, an example of which is shown below.
\begin{lstlisting}
640 480
504.261 503.905
352.457 272.202

640 480
573.71 574.394
346.471 249.031

0.999749 0.00518867 0.0217975 0.0243073
-0.0051649 0.999986 -0.0011465 -0.000166518
-0.0218031 0.00103363 0.999762 0.0151706

1135.09 0.0819141
\end{lstlisting}
This includes (i) for the each camera (RGB or depth) the image size, focal length and principal point (in pixels), as outputted by the Matlab Calibration Toolbox \cite{bouguet_2004} (ii) the extrinsic matrix mapping depth into RGB coordinates obtained and (iii) the calibration of the depth.

We also provide several data sources in the \texttt{InputSource} namespace, examples of those are OpenNI and image files. We tested OpenNI with PrimeSense-based long range depth camera, the Kinect Sensor, RealSense cameras and the Structure Sensor; whereas images need to be in the PPM/PGM format. When only depth is provided (such as e.g. by the Structure Sensor) InfiniTAM will only function with the ICP depth-based tracker.

All library settings are defined inside the \texttt{ITMLibSettings} class, and are:
\begin{lstlisting}
class ITMLibSettings
{
public:
/// The device used to run the DeviceAgnostic code
typedef enum {
DEVICE_CPU,
DEVICE_CUDA,
DEVICE_METAL
} DeviceType;

typedef enum
{
FAILUREMODE_RELOCALISE,
FAILUREMODE_IGNORE,
FAILUREMODE_STOP_INTEGRATION
} FailureMode;

typedef enum
{
SWAPPINGMODE_DISABLED,
SWAPPINGMODE_ENABLED,
SWAPPINGMODE_DELETE
} SwappingMode;

typedef enum
{
LIBMODE_BASIC,
LIBMODE_LOOPCLOSURE
}LibMode;

/// Select the type of device to use.
DeviceType deviceType;

/// Whether or not to perform full raycast for every frame or forward project points from one frame to the next according to the estimate camera pose.
bool useApproximateRaycast;

/// Whether or not to apply a bilateral filtering step on the processed depth images.
bool useBilateralFilter;

/// For ITMColorTracker: skip every other point in energy function evaluation.
bool skipPoints;

/// Whether or not to create the meshing engine thus allowing the saving of meshes from the reconstruction.
bool createMeshingEngine;

/// Sets the behaviour of the system after tracking failure.
FailureMode behaviourOnFailure;

/// Sets the swapping mode.
SwappingMode swappingMode;

/// Whether or not to enable the multi-scene pipeline and loop closure detection.
LibMode libMode;

/// A string specifying the tracker to use and its associated parameters.
const char *trackerConfig;

/// Further, scene specific parameters such as voxel size.
ITMSceneParams sceneParams;

/// And surfel scene specific parameters as well.
ITMSurfelSceneParams surfelSceneParams;
\end{lstlisting}

\section{New Features in InfiniTAM v3}
\label{s:newfeatures}

\subsection{Random Ferns Relocaliser}
\label{s:relocaliser}

One of the key new features that has been added to InfiniTAM since the previous version is an implementation \cite{kahler_eccv_2016} of Glocker et al.'s keyframe-based random ferns camera relocaliser \cite{glocker_tvcg_2015}. This can be used both to relocalise the camera when tracking fails, and to detect loop closures when aiming to construct a globally-consistent scene (see Subsection~\ref{s:loopclosure}). Relocalisation using random ferns is comparatively straightforward. We provide a brief summary here that broadly follows the presentation in \cite{glocker_tvcg_2015}. Further details can be found in the original paper.

At the core of the approach is a method for encoding an RGB-D image $I$ as a set of $m$ binary code blocks, each of length $n$. Each of the $m$ code blocks is obtained by applying a random fern to $I$, where a fern is a set of $n$ binary feature tests on the image, each yielding either $0$ or $1$. Letting $b_{F_k}^I \in \mathbb{B}^n$ denote the $n$-bit binary code resulting from applying fern $F_k$ to $I$, and $b_C^I \in \mathbb{B}^{mn}$ denote the result of concatenating all $m$ such binary codes for $I$, it is possible to define a dissimilarity measure between two different images $I$ and $J$ as the block-wise Hamming distance between $b_C^I$ and $b_C^J$. In other words,
\begin{equation}
\textup{BlockHD}(b_C^I, b_C^J) = \frac{1}{m} \sum_{k=1}^m (b_{F_k}^I \equiv b_{F_k}^J),
\end{equation}
where $(b_{F_k}^I \equiv b_{F_k}^J)$ is $0$ if the two code blocks are identical, and $1$ otherwise.

Given this approach to image encoding, the idea behind the relocaliser is conceptually to learn a lookup table from encodings of keyframe images to their known camera poses (e.g.\ as obtained during a successful initial tracking phase). Relocalisation can then be attempted by finding the nearest neighbour(s) of the encoding of the current camera input image in this table, and trying to use their recorded pose(s) to restart tracking. In practice, a slightly more sophisticated scheme is used. Instead of maintaining a lookup table from encodings to known camera poses, the approach maintains (i) a global lookup table $P$ that maps keyframe IDs to known camera poses, and (ii) a set of $m$ code tables (one per fern) that map each of the $2^n$ possible binary codes resulting from applying a fern to an image to the IDs of the keyframes that achieve that binary code.

This layout makes it easy to find, for each incoming camera input image $I$, the most similar keyframe(s) currently stored in the relocaliser: specifically, it suffices to compute the $m$ code blocks for $I$ and use the code tables to look up all the keyframes with which $I$ shares at least one code block in common. The similarity of each such keyframe to $I$ can then be trivially computed, allowing the keyframes to be ranked in descending order of similarity. During the training phase, this is used to decide whether to add a new keyframe to the relocaliser. If the most similar keyframe currently stored is sufficiently dissimilar to the current camera input image $I$, a new entry for $I$ is added to the global lookup table and the code tables are updated based on $I$'s code blocks. During the relocalisation phase, the nearest keyframe(s) to the camera input image are simply looked up, and their poses are used to try to restart tracking.

The implementation of this approach in InfiniTAM can be found in the \texttt{FernRelocLib} library. The ferns to be applied to an image are stored in a \texttt{FernConservatory}. The \texttt{computeCode} function in \texttt{FernConservatory} computes the actual binary codes. The global lookup table is implemented in the \texttt{PoseDatabase} class, and the code tables are implemented in the \texttt{RelocDatabase} class. Finally, everything is tied together by the top-level \texttt{Relocaliser} class, which provides the public interface to the relocaliser. For readers who want to better understand how the code works, a good place to start is the \texttt{ProcessFrame} function in \texttt{Relocaliser}.

\paragraph{Limitations.} The random ferns approach to relocalisation has a number of advantages: it is easy to implement, fast to run, and performs well when relocalising from poses that correspond to stored keyframes. However, because it is a keyframe-based approach, it tends to relocalise poorly from novel poses. If desired, better relocalisation from novel poses can be obtained using correspondence-based methods, e.g.\ the scene coordinate regression forests of Shotton et al.\ \cite{shotton_cvpr_2013}. The original implementation of this approach required extensive pre-training on the scene of interest, making it impractical for online relocalisation, but recent work by Cavallari et al.\ \cite{cavallari_cvpr_2017} has removed this limitation.

\subsection{Globally-Consistent Reconstruction}
\label{s:loopclosure}

A major change to InfiniTAM since the previous version is the introduction of support for globally-consistent reconstruction \cite{kahler_eccv_2016}. Central to this is the division of the scene into multiple (rigid) submaps, whose poses relative to each other can be optimised in the background to achieve a consistent global map. The system maintains a list of submaps, which it divides into two categories: \emph{active} submaps are tracked against at each frame; \emph{passive} submaps (in which tracking has been lost) are maintained, but not tracked against unless they become active again at some point. One of the active submaps is denoted as the \emph{primary} submap, into which new information from the camera is fused (as in the normal pipeline). Initially, the list contains only a single submap, which is marked as primary. As fusion proceeds, a new submap will be created whenever the camera viewport moves away from the central part of the previous submap.

The optimisation of the relative poses between the different submaps is driven by the accumulation of inter-submap constraints. Constraints between \emph{active} submaps are added during tracking. Additional constraints between the primary submap and passive submaps are added as a result of loop closure detection: when this happens, the passive submaps involved also become active again. In the current implementation \cite{kahler_eccv_2016}, loop closures are detected using the fern-based relocaliser described in Subsection~\ref{s:relocaliser}, although other relocalisers could also be used. Having accumulated these inter-submap constraints, a submap graph optimisation is periodically triggered on a background thread to optimise the relative poses between the submaps. We direct the interested reader to \cite{kahler_eccv_2016} for the details of this optimisation.

The result of this process is a globally-consistent map (`global map') consisting of a collection of submaps whose poses relative to each other have been appropriately optimised. To render the global map, \cite{kahler_eccv_2016} defines a new, combined TSDF $\hat{F}$ that implicitly fuses the individual submaps on-the-fly:
\begin{equation}
\hat{F}(\mathbf{X}) = \sum_i F_w(\mathbf{P}_i \mathbf{X}) F(\mathbf{P}_i \mathbf{X})
\end{equation}
In this, $\mathbf{P}_i$ is the pose of submap $i$, and $F$ and $F_w$ respectively look up the TSDF and weight values in a voxel. Rendering is then performed by raycasting against this combined map. It should be noted that the combined map itself is never explicitly constructed: instead, it acts as a `view' over the individual submaps during rendering. However, the same scheme could be used to explicitly fuse the submaps to construct a single map as the final result if desired.

The implementation of globally-consistent reconstruction in InfiniTAM can largely be found in two parts of the code: \texttt{ITMMultiEngine} class in \texttt{ITMLib}, which contains the pipeline, and the \texttt{MiniSlamGraphLib} library, which handles the pose graph optimisation. The multi-scene pipeline can be enabled in the \texttt{InfiniTAM} application by setting the \texttt{libMode} in the application settings object to \texttt{LIBMODE\_LOOPCLOSURE} instead of \texttt{LIBMODE\_BASIC}. For readers who want to better understand how the code works, a good place to start is the \texttt{ProcessFrame} function in \texttt{ITMMultiEngine}.

\subsection{Surfel-Based Reconstruction}

\begin{figure}[!h]
\centering
\includegraphics[width=.8\linewidth]{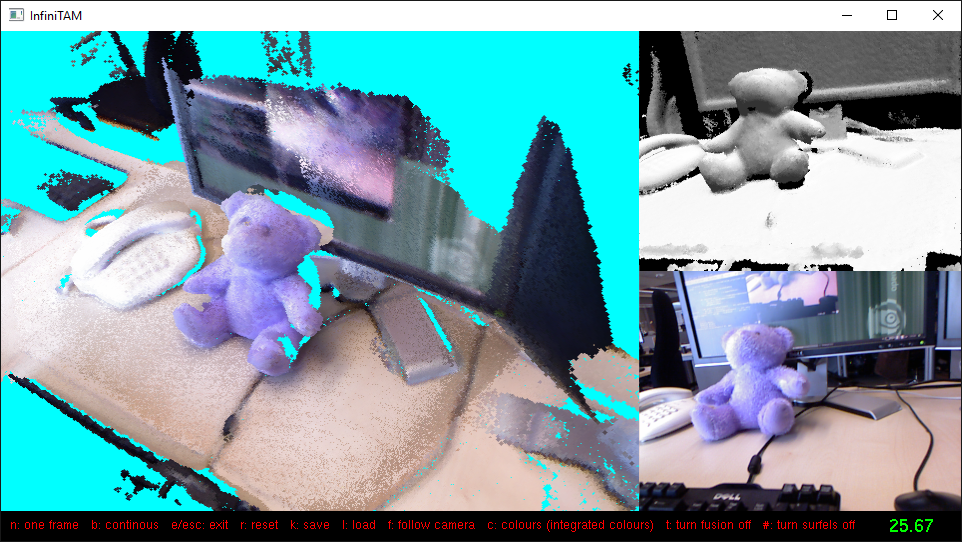}
\caption{The results of running our surfel-based reconstruction engine on InfiniTAM's \emph{Teddy} sequence.}
\label{fig:surfels}
\end{figure}

\noindent In addition to voxel-based reconstruction using TSDFs, InfiniTAM now contains a beta implementation of Keller et al.'s surfel-based reconstruction approach \cite{keller_3dtv_2013}, as shown in Figure~\ref{fig:surfels}.\footnote{We have not yet implemented the hierarchical region growing component of \cite{keller_3dtv_2013}, but everything else is ready to use.} Surfels are essentially just glorified 3D points that have a position $\bar{\mathbf{v}}$, normal $\bar{\mathbf{n}}$, radius $\bar{r}$, confidence count $\bar{c}$ and time stamp $\bar{t}$. At a basic level, Keller's method works by incrementally fusing new points from the live camera input into an existing map of the scene. At each frame, it attempts to find, for each point $\mathbf{v}^g(\mathbf{u})$ in the input with normal $\mathbf{n}^g(\mathbf{u})$, a corresponding surfel in the scene, with position $\bar{\mathbf{v}}_k$, normal $\bar{\mathbf{n}}_k$ and confidence count $\bar{\mathbf{c}}_k$. To do this, it forward projects all the surfels in the scene onto the image plane and writes their indices into the relevant pixels, thus creating an \emph{index map}. Since surfels are circular, their projections on the image plane should really be ellipses, but in practice we follow the common practice of rendering them as circles for simplicity. The corresponding surfel for a point (if any) is then the surfel whose index is written into its pixel in the index map. If a corresponding surfel is found, the point is used to update it via a weighted averaging scheme:
\begin{equation}
\mathbf{\bar{v}}_k \leftarrow \frac{\bar{c}_k \mathbf{\bar{v}}_k + \alpha \mathbf{v}^g(\mathbf{u})}{\bar{c}_k + \alpha}, \;\;\;\; \mathbf{\bar{n}}_k \leftarrow \frac{\bar{c}_k \mathbf{\bar{n}}_k + \alpha \mathbf{n}^g(\mathbf{u})}{\bar{c}_k + \alpha}, \;\;\;\; \bar{c}_k \leftarrow \bar{c}_k + \alpha, \;\;\;\; \bar{t}_k \leftarrow t
\end{equation}
In this, $\alpha$ denotes a measure of sample confidence. If no corresponding surfel is found, a new surfel is added with a confidence count of $\bar{c}_k = \alpha$. Further details can be found in \cite{keller_3dtv_2013}.

The surfel implementation in InfiniTAM mirrors the existing voxel infrastructure. The scene is stored in an instance of the \texttt{ITMSurfelScene} class, which is templated on the surfel type. Two types of surfel are implemented: greyscale surfels in \texttt{ITMSurfel\_grey} and coloured surfels in \texttt{ITMSurfel\_rgb}. Surfel scene reconstruction is implemented in \texttt{ITMSurfelSceneReconstructionEngine} (and its derived classes), and scene visualisation in \texttt{ITMSurfelVisualisationEngine} (and its derived classes). There are also classes such as \texttt{ITMDenseSurfelMapper} and \texttt{ITMSurfelRenderState} that mirror the functionality of their voxel counterparts. For readers who want to better understand how the code works, a good place to start is the \texttt{IntegrateIntoScene} function in \texttt{ITMSurfelSceneReconstructionEngine}, which contains the main surfel reconstruction pipeline. The key bits of code for surfel scene visualisation can found in the shared header for \texttt{ITMSurfelVisualisationEngine}: in particular, readers may want to take a look at the \texttt{update\_depth\_buffer\_for\_surfel} and \texttt{update\_index\_image\_for\_surfel} functions.

\paragraph{Limitations.} Since surfel support in InfiniTAM v3 is still in beta, there are a number of limitations to the existing implementation that readers should note. Firstly, the number of points that a surfel scene can contain is currently limited to $5$ million (see the \texttt{MAX\_SURFEL\_COUNT} constant if you want to change this). This isn't a huge problem in practice given the Keller method's support for surfel removal and surfel merging, but it's worth knowing about. Secondly, our surfel reconstruction implementation is currently significantly less optimised than our voxel one: it's still real-time, but needs a fairly good GPU to work well. Finally, there is currently no support for generating meshes from a surfel scene, and it is also currently not possible to use surfel scenes with our loop closure implementation. Both of these may change in the future, if we find the time.

\bibliographystyle{plain}
\bibliography{InfiniTAM}

\end{document}